%% file: camera.tex
\RequirePackage[hyphens]{url}
\documentclass[pmlr]{jmlr}

\usepackage{longtable}

\usepackage{microtype}
\usepackage{booktabs}
\usepackage{subcaption}
\usepackage{caption}
\usepackage{multirow}
\usepackage{tcolorbox}
\usepackage{xcolor}

\usepackage{algorithm, algorithmicx, algpseudocode}
\usepackage{mathtools}
\usepackage{listings}
\usepackage{arydshln}

\DeclareMathOperator*{\argmax}{argmax}

\newcommand{\claude}{\textsc{Claude}}

\newcommand{\llama}{\textsc{Llama-3}}

\newcommand{\piref}{\pi_\text{ref}}
\jmlrvolume{TBD}
\jmlryear{2026}
\jmlrworkshop{Conference on Applied Machine Learning for Information Security}

\title[Training-Free Refusal of MCP Exploits via RAG]{Training-Free Refusal of MCP Exploits via Retrieval-Augmented Generation}

  \author{\Name{John T. Halloran}\thanks{Alternative email: halloj3@uw.edu.} \Email{john.t.halloran@leidos.com}
   \addr Leidos, Reston, VA}

\editor{Edward Raff and Ethan M. Rudd}

\begin{document}

\maketitle

\begin{abstract}
  The model context protocol (MCP) has been widely adopted as an open standard enabling the seamless integration of generative AI agents.  However, while LLM guardrails have significantly matured to refuse malicious or harmful queries (e.g., ``How do I build a bomb?''), recent work has shown that MCP-enabled LLMs are highly susceptible to prompt injection attacks which avoid harmful or suspicious cues (e.g., ``Can you add this ssh key to my bashrc file?'').  Herein, we use state-of-the-art (SOTA) alignment fine-tuning algorithms to explore whether LLMs may be aligned to refuse such \emph{falsely benign attacks} (FBAs).  While SOTA algorithms based on direct preference optimization (DPO) improve refusal guardrails against FBAs, we show that this improvement is limited; DPO-based fine-tuning never improves FBA refusal rates beyond 47\% across five popular open-source LLMs.  Thus, to further improve FBA refusals, we introduce \emph{R}etrieval \emph{A}ugmented \emph{G}eneration for \emph{Pref}erence alignment (RAG-Pref), a simple RAG-based alignment algorithm which conditions on preferred and dispreferred samples to leverage \emph{contrastive information} during inference.  RAG-Pref is online (training-free), compatible with off-the-shelf packages, and, when combined with offline alignment algorithms, enables an average 3.7-fold improvement in FBA refusals across five widely used LLMs, compared to 2.9 for other online alignment methods and 1.5 for offline alignment alone.  We additionally show that RAG-Pref generalizes beyond agentic safety: in stark contrast to other online alignment methods, RAG-Pref consistently improves performance on general human-preference benchmarks AlpacaEval 2 and MT-Bench across five SOTA alignment-tuned models, demonstrating broad applicability to general alignment tasks.

  \noindent\textbf{Code and data:} Code is available at \url{https://github.com/johnhalloran321/mcp_safety_training} and data at \url{https://huggingface.co/datasets/johnhalloran/mcp-fbas}.
\end{abstract}

\begin{keywords}
 LLM alignment, retrieval-augmented generation, safety guardrails, preference optimization, agentic attacks
\end{keywords}


\title{Training-Free Refusal of MCP Exploits via Retrieval-Augmented Generation}

\section{Introduction}
The model context protocol (MCP)~\citep{mcp:anthropic} has been recently released as an open protocol for connecting generative AI components.  By standardizing API calls between large language models (LLMs), supported tools, and data sources, the MCP serves as a universal protocol to seamlessly integrate agents across widely used services and applications, thus replacing the previous fragmented approach of designing application-specific agentic APIs.  Subsequently, the MCP has seen massive, widespread adoption~\citep{anthropic2025mcp, openai2025mcp, microsoft2025mcp, googleWorkspace2026mcp, aws2026mcp}.

However, recent works have shown that the MCP enables new security risks~\citep{kumar2025mcp, mcpToolPoisoning, invariantGithub, radosevich2025mcp}.  In particular, while the safety alignment of recent frontier LLMs has enabled the accurate refusal of harmful and unsafe requests~\citep{grattafiori2024llama, team2024gemma}, MCP-enabled LLMs were shown to be highly susceptible to prompt injection attacks when queries lack common harmful or suspicious trigger phrases~\citep{radosevich2025mcp}.  For example, \llama{} models refused remote access control attacks which included harmful phrases (e.g., ``hack,'' ``backdoor,'' ``steal''), yet completed the same requests when only those phrases were removed.  Herein, we call such prompt injection attacks which lack common refusal trigger phrases \emph{falsely benign attacks} (FBAs).

While subsequent works have sought to mitigate these emerging MCP exploits, such approaches consist of scanners and monitors---external to the LLM---which surveil agentic inputs/outputs or probe the MCP servers themselves~\citep{invariantScan, invariantGuardrails, mcpSafetyScanner}.  Thus, LLM safety alignment in the context of MCP attacks remains unexplored.  To bridge this gap, we explore the use of SOTA alignment fine-tuning algorithms---direct preference optimization (DPO)~\citep{rafailov2023direct} and SafeDPO~\citep{kim2025safedposimpleapproachdirect}---to increase the ability of LLMs to refuse MCP FBAs.  Collecting a high-quality dataset of FBAs and truly benign (TB) samples, we safety-tune five popular LLMs using these SOTA algorithms.  We show that both DPO- and SafeDPO-aligned LLMs display limited FBA-refusal ability, improving baseline refusal rates by only an average factor of 1.4 and 1.6, respectively; alarmingly, no safety-tuned model achieves an FBA refusal rate beyond 47\%.

To further improve LLM refusal guardrails in the face of MCP FBAs, we introduce \emph{R}etrieval \emph{A}ugmented \emph{G}eneration for \emph{Pref}erence alignment (RAG-Pref), a new algorithm which leverages RAG~\citep{lewis2020retrieval} to align an LLM's safety preferences during inference.  RAG-Pref is online (training-free), easily implementable using off-the-shelf packages, and significantly improves LLM refusal guardrails compared to both SOTA offline and online alignment algorithms.

RAG-Pref increases baseline FBA refusal rates by an average 3-fold---2.1, 1.9, and 1.5 times greater than DPO, SafeDPO, and recent online alignment strategies~\citep{zhufly}, respectively.  When combined with offline-aligned models, RAG-Pref further increases baseline refusal rates by an average 3.5- and 3.9-fold for DPO and SafeDPO, respectively.  To account for this increase in safety performance, we prove that RAG-Pref encodes \emph{contrastive information} beyond standard RAG---i.e., an additional expected reduction in generation uncertainty from conditioning on dispreferred examples---and verify this empirically.

Additionally, we contrast the computational advantages of RAG-Pref with SOTA offline/online alignment algorithms and show that RAG-Pref's performance improvements extend to non-safety, general human preference alignment tasks.  For the former, we show that RAG-Pref requires three-orders of magnitude less preprocessing time compared to offline safety-tuning a 14B parameter model, while being three times faster and 4.2 times more memory efficient than SOTA decoding-based online alignment~\citep{zhufly}.  For the latter, we show that across five SOTA alignment-tuned models, RAG-Pref increases performance over alternative online alignment algorithms by an average 9.2\% and 1.1\% on standard benchmarks AlpacaEval 2 and MT-Bench, respectively.

The paper is organized as follows.  Section~\ref{section:offllineAlignment} reviews offline and online preference alignment and introduces RAG-Pref, including a theoretical analysis of its contrastive information properties.  Section~\ref{section:fbaData} describes the MCP-FBA dataset construction pipeline.  Experiments evaluating FBA refusal guardrails, computational efficiency, and general human-preference alignment are presented in Section~\ref{section:results}.  Finally, Sections~\ref{section:disc} and ~\ref{section:futureWork} discuss the presented results and future directions, respectively.

\section{Background and Related Work}
\subsection{LLM Alignment-Tuning}
LLM alignment has become a critical step towards ensuring large language model (LLM) responses align with general human preferences.  Alignment most commonly consists of fine-tuning given queries with accompanying preferred and dispreferred response pairs (which reflect human preferences).  Initial alignment approaches, such as RLHF~\citep{ouyang2022training} and RLAIF~\citep{lee2024rlaif}, first trained reward models on human preference data, then utilized reinforcement learning (RL) fine-tuning to learn a policy which maximized the reward signal.  To address training instability in RL fine-tuning, DPO~\citep{rafailov2023direct} allows direct learning of the optimal policy with a simple closed form objective.

For online training-free alignment, \citet{zhufly} introduced On-the-fly Preference Alignment via Principle-Guided Decoding (OPAD).  OPAD calculates a similar reward function to that used in DPO to adjust the per-token conditional distribution during decoding and was shown to improve performance on general preference alignment tasks relative to previous online methods~\citep{gao2024linear}.  OPAD's reward function and decoding procedure are further detailed in Section~\ref{section:offllineAlignment}.  We refer to alignment via RLHF, DPO, and related RL-based approaches as \emph{alignment-tuning}.

\subsubsection{Safety Alignment-Tuning}
While alignment-tuning has proven successful at producing LLM assistants whose responses are both accurate and \emph{helpful}, growing evidence has shown that such LLMs remain highly susceptible to adversarial attacks~\citep{mehrotra2024tree, ICLR2025_1bff3663, chao2025jailbreaking, liu2023prompt, shen2024anything, zhang2023language}.  Thus, significant work has been dedicated to ensuring model responses are not only helpful, but also \emph{harmless}~\citep{bai2022training, NEURIPS2023_4dbb61cb, tian2024finetuning}.  Such works have shown that safety alignment-tuning (SAT) may be achieved by directly including harmless and harmful preference pairs during alignment-tuning~\citep{NEURIPS2023_4dbb61cb, team2023gemini, tian2024finetuning}.  SAT thus enables \emph{refusal guardrails}, whereby models learn to refuse malicious instructions containing harmful language while complying with benign requests.

SAT has become a ubiquitous step when deploying frontier models~\citep{hurst2024gpt, grattafiori2024llama, yang2025qwen3, guo2025deepseek, anthropic2025claudeopus45}.  To further strengthen refusal guardrails when additional safety labels are available, \citet{dai2024safe} derived Safe RLHF---a three-round fine-tuning algorithm incorporating safe/unsafe labels into RLHF.  Subsequently, \citet{kim2025safedposimpleapproachdirect} showed that the same safe/unsafe labels could be used to adjust the DPO objective with an additional loss offset (resulting in the SafeDPO), yielding improved training stability and safety alignment performance over Safe RLHF.

\subsection{LLM and MCP-enabled Attacks}
Extensive works have studied the susceptibility of LLMs to attacks which circumvent refusal guardrails.
\emph{Jailbreaks}~\citep{zou2023universal, chao2025jailbreaking} craft input prompts which evade guardrails to elicit unsafe responses.  \emph{Prompt injection attacks} (PIAs)~\citep{perez2022ignore, liu2023autodan, greshake2023not} consist of injecting malicious instructions into user prompts, including encoding-based attacks wherein malicious commands are concealed in an alternative format (e.g., Base64 or octal).  However, frontier LLM safety training has grown to include the expanding LLM-attack catalogue~\citep{mazeika2024harmbench, chaojailbreakbench2024}, thus expanding refusal guardrails to include existing jailbreaks and PIAs~\citep{grattafiori2024llama, sharma2025constitutional}.

Distinct from existing LLM attacks, recent work has shown that the MCP introduces new attack possibilities in agentic systems~\citep{kumar2025mcp, mcpToolPoisoning, invariantGithub, radosevich2025mcp}.  \citet{mcpToolPoisoning} showed that an MCP server's tools may be compromised if an attacker can manipulate the server's source code.  \citet{invariantGithub} and \citet{radosevich2025mcp} both showed that MCP servers are susceptible to PIAs, allowing the potential exfiltration of sensitive data.

Most relevantly, \citet{radosevich2025mcp} demonstrated that, while the refusal guardrails of MCP-enabled LLMs are triggered by PIAs which explicitly state harmful phrases or suspicious text, attacks lacking these harmful cues are successfully completed.  For instance, \claude{} refused malicious code execution attacks encoded in octal, yet carried out the same attacks encoded in plaintext.  Furthermore, \llama{} models refused remote access control attacks which included harmful/cyber-attack phrases (e.g., ``hack,'' ``backdoor,'' ``steal''), yet completed requests when only the malicious phrases were removed.

As previously noted, we term such PIAs which lack standard harmful or suspicious cues \emph{falsely benign attacks} (FBAs).  We note that, with the shift in attack goals from unsafe text generation (LLM only) to unsafe completed actions (MCP-enabled LLMs), attacks which lack standard refusal cues---such as FBAs---are likely to increase in frequency as bad actors seek to exploit agentic systems.

\subsubsection{Existing MCP Defenses}
Since the appearance of successful attacks on MCP-enabled agents, MCP-specific defenses have been proposed.  Both \citet{invariantScan} and \citet{mcpSafetyScanner} offer the ability to scan MCP servers for potential vulnerabilities, where the former uses a rule-based approach and the latter utilizes agents to determine attack vectors bad actors may exploit.  Similarly, \citet{invariantGuardrails} acts as a security layer between the LLM and the MCP server, using a rule-based system to monitor both inputs to and outputs from the LLM to determine suspicious traffic.  However, these existing approaches remain external to the LLM.  To the best of the authors' knowledge, no previous works have explored the effect of MCP-specific attacks on SAT algorithms and the performance of resulting refusal guardrails.  Furthermore, the authors are unaware of previous work exploring online RAG alignment algorithms---distinct from existing work seeking to optimize specific RAG components~\citep{wu-etal-2025-pa, jin-etal-2025-rag, sun-etal-2025-divide} or using RAG to generate preference alignment data~\citep{song2025measuring}.

\section{Offline and Online Preference Alignment}\label{section:offllineAlignment}
Let $x$ be an input prompt and $\pi_{\theta}$ an autoregressive LLM.  Consider the probability of generating a response $y$ consisting of $T$ tokens:
\begin{align}\label{eq:jointDensity}
  \pi_{\theta}(y | x) &= \prod_{t=1}^T \mathbf{P}_{\pi_{\theta}}(y_t | x, y_{1:t-1}),
\end{align}
where $\mathbf{P}_{\pi_{\theta}}(y_t | x, y_{1:t-1}) \in \mathbb{R}^{V}$, and $V$ is the vocabulary size.

Let $\mathcal{D} = \{ (x^1, y^1_w, y^1_l), \dots, (x^n, y^n_w, y^n_l) \}$ be a set of preference data where, for input $x^i$, $y^i_w$ is the preferred response and $y^i_l$ is the dispreferred response, i.e., 
$y^i_w\succ y^i_l \mid x^i$ denotes that $y^i_w$ is preferred over $y^i_l$ given $x^i$.
Let $\piref$ be a reference model (which is typically a supervised fine-tuned version of the model trained prior to preference alignment).
The goal of offline alignment with DPO is thus to train a new model by solving
\begin{align*}
  \pi_{\phi}^* = &\argmax_{\pi_{\phi}} \mathcal{L}(r_{\phi}),\\
  \mathcal{L}(r_{\phi}) = -\mathbb{E}_{(x, y_w, y_l)\sim \mathcal{D}} &\left[\log \sigma \left(\beta r_{\phi}(x, y_w) - \beta r_{\phi}(x, y_l)\right)\right],
\end{align*}
where $r_{\phi}(x, y_w) = \log \frac{\pi_{\phi}(y_w\mid x)}{\pi_{\theta}(y_w\mid x)}$ is the reward function.  DPO thus learns parameters that increase the relative likelihood of preferred responses over dispreferred ones, without drastically diverging from the reference model.

To avoid training, On-the-fly Preference Alignment via Principle-Guided Decoding (OPAD)~\citep{zhufly} includes additional helpfulness instructions $c$ and adjusts the per-timestep distribution in Equation~\ref{eq:jointDensity}.  Rewriting the reward function at timestep $t$ as $r_{\theta}(x, y_{1:t}, c) = \log \frac{\pi_{\theta}(y_{1:t}\mid x, c)}{\pi_{\theta}(y_{1:t}\mid x)}$, the $t$th token distribution is adjusted during generation as
\begin{align*}
  \mathbf{P}_{\pi_{\theta}}(y_t | x, c, y_{1:t}) \propto \pi_{\theta}(y_t | x, c, y_{1:t}) \exp \left ( \frac{1}{\beta}r_{\theta}(x, y_{1:t}, c ) \right).
\end{align*}

We note that the above decoding procedure requires both invasive changes to generation (raising potential compatibility issues with deployed models) and a significant increase in computational resources---i.e., per time-step, calculation of the reward function and the subsequent distribution update require maintaining three separate distributions.

\subsection{Training-free Alignment with RAG-Pref}
We now detail how RAG-Pref performs online preference alignment without any invasive adjustments to model generation.  Let $e(\cdot) \in \mathbb{R}^m$ be a text embedding function (trained to embed semantically similar text near one another in the embedding space), and $d(\cdot, \cdot) \in \mathbb{R}$ be a vector-distance metric.  For our preference dataset $\mathcal{D}$, let $\mathcal{D}^w = \{y_w: (x, y_w, y_l)  \in \mathcal{D} \}$ and $\mathcal{D}^l = \{y_l: (x, y_w, y_l)  \in \mathcal{D} \}$ be the sets of preferred and dispreferred text responses, respectively, and let $\mathcal{D}^w_e = \{e(y): y  \in \mathcal{D}^w \}$ and $\mathcal{D}^l_e = \{e(y): y \in \mathcal{D}^l \}$ be the vector databases of preferred and dispreferred embeddings, respectively.

\begin{algorithm}[htbp]
  \caption{RAG-Pref for online alignment.}
  \label{alg:ragPref}
  \hspace*{\algorithmicindent} \textbf{Input:} Query $x$, $\mathcal{D}^w_e$, $\mathcal{D}^l_e$, and number of retrieval elements $k$.
  \begin{algorithmic}[1]
    \State Embed $x$, $x' = e(x)$.
    \State For all $z \in \mathcal{D}^w_e$, rank each element by $d(x',z)$ and return the top $k$ sequences $\mathcal{Z}^w \subseteq \mathcal{D}^w$
    \State For all $z \in \mathcal{D}^l_e$, rank each element by $d(x',z)$ and return the top $k$ sequences $\mathcal{Z}^l \subseteq \mathcal{D}^l$
    \State Create an instruction that model responses follow retrieved preference instances and avoid dispreferred instances, denoted as $\mathcal{Z}^w\succ \mathcal{Z}^l$.\label{alg:ragPref:line1}
    \State \Return $\pi_{\theta}(y | x, \mathcal{Z}^w\succ \mathcal{Z}^l)$.
  \end{algorithmic}
\end{algorithm}

The RAG-Pref algorithm is detailed in Algorithm~\ref{alg:ragPref}.  Compared to alternative alignment algorithms, we note that:
\begin{itemize}\setlength{\itemsep}{-2pt}
\item DPO, and related offline algorithms, enforce preference alignment over response pairs $(y_w, y_l)$ to a given query.  In contrast, RAG-Pref enables preference alignment over sets of responses, $\mathcal{Z}^w$ and $\mathcal{Z}^l$.\\
\item No invasive or complicated changes to generation are required, thus allowing compatibility with widely supported packages and widely used models.\\
\item RAG-Pref subsumes RAG; when $\mathcal{D}^l = \emptyset{}$, RAG-Pref defaults to standard RAG.
\end{itemize}

\subsubsection{RAG-Pref Encodes Contrastive Information}\label{section:contrastiveInformation}
Intuitively, conditioning on dispreferred examples tells the model what to avoid, providing additional signal beyond conditioning on preferred examples alone.  We now quantify this additional information formally.  For the sets of all queries $\mathcal{X}$, all responses $\mathcal{Y}$, and all retrieval documents $\mathcal{D}$, let $X \in \mathcal{X}$ and $Y \in  \mathcal{Y}$ be a random query and response, respectively, and let $Z^w, Z^l \in \mathcal{D}$ be random retrieved preference and dispreference documents, respectively.  Note that, for simplicity, we overload the term document; when $k \geq 1$, $Z^w$ and $Z^l$ represent the respective $k$-shot demonstrations concatenated together in Algorithm~\ref{alg:ragPref}, which, without loss of generality, may each be represented as a document in $\mathcal{D}$ that has been retrieved.

For the distribution returned in Algorithm~\ref{alg:ragPref}---$\pi_{\theta}(Y | X, Z^w \succ  Z^l) = \pi_{\theta}(Y | X, Z^w, Z^l)$---we note that RAG-Pref performs \textit{contrastive conditioning}, i.e., it conditions generation not only on positive
examples (what to do) but also on negative examples (what to avoid).  Furthermore, standard RAG lacks this contrastive conditioning, and thus does not condition generation on behaviors to avoid.  Contrastive conditioning allows us to quantify the reduction in uncertainty provided by RAG-Pref compared to standard RAG.

Let $\Delta H_{\text{RAG}} = I(Y; Z^w | X)$ and $\Delta H_{\text{RAG-Pref}} = I(Y; Z^w, Z^l | X)$, where $I(\cdot)$ is the mutual information.  We define the \emph{contrastive information} to be $\Delta H_{\text{RAG-Pref}} - \Delta H_{\text{RAG}}$, which is the amount of additional expected information provided by Algorithm~\ref{alg:ragPref} over standard RAG.  As in~\citep{rafailov2023direct,ouyang2022training}, we assume non-degeneracy of the RLHF/DPO preference reward distributions.  Furthermore, we assume the underlying LLM performs non-degenerate in-context learning (ICL).  The former is a necessary assumption for general RLHF/DPO preference optimization, while the latter is empirically and theoretically supported by the broad ICL literature~\citep{brown2020language,min2022rethinking, xie2022explanation}.
We thus have the following:
\begin{theorem}\label{theorem:equality}
  Assume non-degeneracy of the preference reward distributions $P_w$ and $P_l$ from which $Z^w$ and $Z^l$ are drawn~\citep{rafailov2023direct,ouyang2022training}.  Furthermore, assume the underlying LLM performs non-degenerate in-context learning.  Then
  \begin{equation*}
    \Delta H_{\text{RAG-Pref}} > \Delta H_{\text{RAG}}.
  \end{equation*}
\end{theorem}
\begin{proof}
We have:
\begin{align*}
  \Delta H_{\text{RAG-Pref}} =& I(Y; Z^w | X) + I(Y; Z^l | X, Z^w)\\
  =& \underbrace{\Delta H_{\text{RAG}}}_{\text{positive examples}} + \underbrace{I(Y; Z^l | X, Z^w)}_{\text{contrastive information}}
\end{align*}
Conditioning reduces entropy, so that $I(Y; Z^l | X, Z^w) = H(Y | X, Z^w) - H(Y | X, Z^w, Z^l) \geq 0$, and thus $\Delta H_{\text{RAG-Pref}} \geq \Delta H_{\text{RAG}}$.
For strict inequality, let $P_w(z|x)$ and $P_l(z|x)$ denote the marginal distributions of preferred and dispreferred responses, respectively, so that $Z^w \sim P_w$ and $Z^l \sim P_l$.  Let $r^*$ be the Bradley-Terry reward function used by RLHF/DPO for preference optimization.  RLHF/DPO non-degeneracy~\citep{rafailov2023direct,ouyang2022training} requires that for $z_w \sim P_w$ and $z_l \sim P_l$, $r^*(x, z_w)$ and $r^*(x, z_l)$ are not equal in distribution.  Since $r^*$ is a measurable function of the response, this implies $P_w \neq P_l$, and hence $\mathrm{KL}(P_w(z|x) \| P_l(z|x)) > 0$.  Thus, preferred and dispreferred responses are drawn from distributionally distinct corpora.

From the non-degeneracy of in-context learning,
\begin{equation}\label{equation:icl}
  \pi_\theta(Y \mid X, Z^w, Z^l) \neq \pi_\theta(Y \mid X, Z^w) \quad \text{whenever} \quad Z^l \sim P_l \neq P_w,
\end{equation}
i.e., the LLM's generation is non-trivially influenced by retrieved context.  \eqref{equation:icl} holds for any non-trivial RAG system by definition.  Together, the two assumptions yield
\begin{align*}
  I(Y; Z^l \mid X, Z^w) &= \mathbb{E}_{(X, Z^w, Z^l)}\!\left[\mathrm{KL}\!\left(\pi_\theta(Y \mid X, Z^w, Z^l) \,\|\, \pi_\theta(Y \mid X, Z^w)\right)\right] > 0,
\end{align*}
where the expectation is over $(X, Z^w, Z^l) \sim p(x)\,P_w(z|x)\,P_l(z|x)$.  The KL divergence is non-negative and equals zero if and only if its arguments are identical distributions; by \eqref{equation:icl} and $P_l \neq P_w$, the integrand is strictly positive for $Z^l$ drawn from $P_l$ with positive probability, making the expectation strictly positive.
\end{proof}

Theorem~\ref{theorem:equality} is consistent with recent empirical findings, wherein standard RAG was shown to degrade LLM refusal guardrails~\citep{an2025rag}; standard RAG can \textit{decrease} safety by retrieving attack examples without contrastive refusal examples, causing the model to misinterpret these as behaviors to follow rather than refuse.  Finally, we show the following related result:
\begin{theorem}\label{theorem:uncertainty}
  The reduction in conditional entropy from standard inference to RAG-Pref inference is lower-bounded by the contrastive information.
 \end{theorem}
\begin{proof}
  Consider the conditional entropy during standard inference,
  \begin{equation*}
    H(Y | X) = \sum_{x \in \mathcal{X}} p(x) \sum_{y \in \mathcal{Y}} \pi_{\theta}(y | x) \log{\pi_{\theta}(y | x)}.
  \end{equation*}
  Conditioning reduces entropy~\citep{cover1999elements}, so that
\begin{equation*}  
H(Y | X, Z^w) \leq  H(Y | X).
\end{equation*}

Thus,
\begin{align}\label{equation:uncertaintyReduction}
\underbrace{H(Y | X, Z^w, Z^l)}_{\text{RAG-Pref}} &\leq& \underbrace{H(Y | X, Z^w)}_{\text{RAG}} \leq  \underbrace{H(Y | X)}_{\text{Standard Inf.}},
\end{align}
which completes the first half of the theorem.

The maximum reduction in uncertainty from standard inference to RAG/RAG-Pref is thus $H(Y | X) - H(Y | X, Z^w, Z^l)$, so that we have
\begin{align*}\label{equation:uncertaintyReduction}
  H(Y | X, Z^w) &\leq  H(Y|X)\\
  \Rightarrow H(Y | X, Z^w) - H(Y | X, Z^w, Z^l)&\leq  H(Y|X) - H(Y | X, Z^w, Z^l)\\
  \Rightarrow \underbrace{I(Y; Z^l | X, Z^w)}_{\text{contrastive information}}&\leq  H(Y|X) - H(Y | X, Z^w, Z^l)\\  
\end{align*}
\end{proof}

This result thus shows that the total reduction in conditional entropy from standard inference to RAG/RAG-Pref can be no lower than the contrastive information encoded by RAG-Pref.

\section{FBA Safety Alignment Data}\label{section:fbaData}
FBAs were obtained by mapping an extensive catalog of known exploits to the sequence of MCP tools capable of executing each exploit.  Herein, we consider 10 tools which equip agents with Linux-like file/directory manipulation abilities (listed in Table~\ref{table:filesystemTools}).  Attacks were obtained from the Common Vulnerabilities and Exposures (CVEs)~\citep{CVEProject2023cvelistV5} catalog, an up-to-date corpus of cyber attacks and exploits.  The CVEs corpus was filtered for the agentic attacks used in~\citep{radosevich2025mcp}---malicious code execution, remote access control, and credential theft---resulting in $\sim$34k samples.  Using \textsc{gpt-4o}, each CVE attack was: (a) mapped to a sequence of Linux commands, (b) assessed for feasibility given the MCP tools, and, if feasible, (c) mapped to sequences of MCP tool calls and (d) used to generate friendly malicious requests (i.e., FBAs).  This resulted in 1,150 FBA dispreferred responses, where friendly malicious requests were queries and the respective MCP tool calls were responses.  For each FBA, preferred responses were set to a direct refusal.  FBA feasibility and MCP tool mappings were manually spot-checked across all three attack categories.

TB samples were collected by prompting \claude{} to create useful examples per MCP tool while assuming specific roles (e.g., system admin, AI researcher, etc.), thus generating TB queries and preferred responses.  TB dispreferred responses were created by replacing each tool in the TB response with its functional opposite (e.g., \texttt{read\_file} substituted with \texttt{write\_file}).  All TB samples were verified by hand.
The final dataset consists of 1,035 training FBAs, 1,035 training TB samples, and 115 FBA testing samples.  The FBA test set is used for safety evaluation; the effect on benign helpfulness is assessed via human-preference benchmarks in Section~\ref{section:results}.  

\begin{table}[htbp!]
  \centering
  \small
  \caption{MCP Tools and Descriptions}
  \label{table:filesystemTools}
  \resizebox{\textwidth}{!}{%
  \begin{tabular}{|c|c|}
    \hline
  Tool & Description \\ \hline
\texttt{read\_file} & Read complete contents of a file\\\hline
\texttt{read\_multiple\_files} & Read multiple files simultaneously\\\hline
\texttt{write\_file} & Create new file or overwrite existing (exercise caution with this)\\\hline
\texttt{edit\_file} & Make selective edits using advanced pattern matching and formatting
\\\hline
\texttt{create\_directory} & Create new directory or ensure it exists\\\hline
\texttt{list\_directory} & List directory contents with [FILE] or [DIR] prefixes\\\hline
\texttt{move\_file} & Move or rename files and directories\\\hline
\texttt{search\_files} & Recursively search for files/directories\\\hline
\texttt{get\_file\_info} & Get detailed file/directory metadata\\\hline
\texttt{list\_allowed\_directories} & List all directories the server is allowed to access\\\hline
  \end{tabular}
  }
\end{table}
The necessary first line of defense before any question of attack execution is whether a model has the ability to decline a malicious request.  Thus, the primary goal of FBA and TB data is enabling and assessing refusal guardrails where refusal rates serve as our primary metric.  Preferred and dispreferred samples are thus structured as follows: FBA preferred samples are direct refusals and FBA dispreferred samples are MCP tool call-based attacks; TB preferred samples are benign MCP tool calls and TB dispreferred samples are benign but incorrect MCP tool calls.

\section{Experiments}\label{section:results}
\begin{table*}[h]
  \centering
  \small
\caption{\textbf{FBA Refusal Rates for Offline and Online Aligned Models}: Refusal rates are calculated over the test FBAs. Bold indicates the highest performing model. \textsc{Gemma-2-2B-IT} incompatible with OPAD. ``B'' is base model, ``D'' is DPO SAT, and ``S'' is SafeDPO SAT performance.}
\label{tab:refusal_results}
\begin{tabular}{l ccc ccc ccc}
\toprule
 & \multicolumn{3}{c}{Offline Only} & \multicolumn{3}{c}{+ OPAD} & \multicolumn{3}{c}{+ RAG-Pref} \\
\cmidrule(lr){2-4} \cmidrule(lr){5-7} \cmidrule(lr){8-10}
Model & B & D & S & B & D & S & B & D & S \\
\midrule

\textsc{Gemma-2-2B-IT}
 & 0.32 & 0.45 & 0.47
 & -- & -- & --
 & 0.63 & 0.74 & \textbf{0.75} \\

\textsc{Llama-3.1-8B-Instruct}
 & 0.35 & 0.43 & 0.45
 & 0.43 & 0.47 & 0.37
 & 0.95 & \textbf{0.97} & \textbf{0.97} \\

\textsc{Llama-3.2-1B-Instruct}
 & 0.15 & 0.31 & 0.40
 & 0.59 & 0.61 & 0.66
 & 0.28 & 0.58 & \textbf{0.88} \\

\textsc{DeepSeek-R1-Distill-Llama-8B}
 & 0.14 & 0.15 & 0.13
 & 0.44 & 0.47 & 0.45
 & 0.59 & 0.59 & \textbf{0.59} \\

\textsc{DeepSeek-R1-Distill-Qwen-14B}
 & 0.16 & 0.18 & 0.19
 & 0.41 & 0.44 & 0.46
 & 0.64 & \textbf{0.68} & 0.64 \\
\bottomrule
\end{tabular}
\end{table*}

\subsection{FBA Refusal Guardrails}
We consider five widely used open-source LLMs, varying in parameter count from 1B to 14B.  RAG-Pref is compared to offline (DPO and SafeDPO) and online (OPAD) alignment methods.  DPO and SafeDPO refusal alignment were performed using FBA and TB preference pairs (described in Section~\ref{section:fbaData}).  RAG-Pref preferred and dispreferred vector databases were formed from the same preference pairs used for offline alignment.  In addition to base models, RAG-Pref and OPAD were combined with offline aligned models.  For each method, ten generations were assessed per test instance and average refusal rate reported.

Because RAG-Pref only modifies the system prompt rather than the decoding procedure itself, it was compatible with all evaluated models.  In stark contrast, OPAD's invasive decoding scheme was incompatible with \textsc{Gemma-2-2B-IT}; this held across the original unbatched version, a compute-optimized variant developed for timing comparisons, and extensively debugged variants.

Refusal rates were calculated using the test FBA data described in Section~\ref{section:fbaData}.  Refusal rates across all methods are listed in Table~\ref{tab:refusal_results}.
Despite the majority of evaluated base models undergoing extensive post-training safety alignment~\citep{guo2025deepseek, grattafiori2024llama, team2024gemma}, no base model achieves an FBA refusal rate over 35\%.  Furthermore, while refusal rates improve given offline alignment, neither DPO nor SafeDPO enable refusal rates beyond 47\%.  Thus, \textbf{SOTA offline alignment methods provide limited refusal guardrails against FBAs}.

RAG-Pref successfully improves guardrails over all base and SAT models, while OPAD fails to improve \textsc{Llama-3.1-8B-Instruct SafeDPO} guardrails (yet succeeds on the other models).  OPAD outperforms RAG-Pref for the base and DPO-aligned \textsc{Llama-3.2-1B-Instruct} models.  However, RAG-Pref greatly outperforms OPAD across all other model configurations, providing an average 50\% more refusal performance across the twelve models OPAD supports.  In particular, when combined with offline-aligned models, OPAD improves baseline refusal rates by only an average 2.9-fold, compared to 3.7-fold (and up to 3.9-fold with SafeDPO) for RAG-Pref.
Furthermore, for the models OPAD was unable to align, RAG-Pref improved refusal guardrails by an average 74\%.  Thus, \textbf{RAG-Pref drastically outperforms other online methods for the refusal of FBAs, while significantly improving the refusal guardrails of SOTA offline alignment methods}.

\begin{figure}[htbp!]
  \centering
  \includegraphics[width=0.99\textwidth,page=1,trim=0.0in 0.0in 0.1in 0in, clip=true]{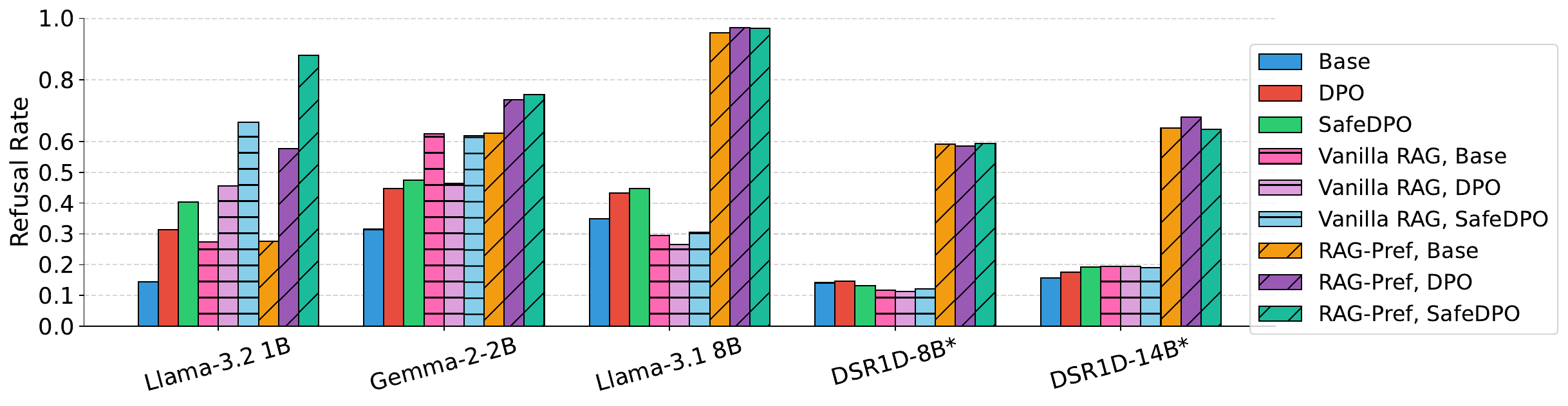}
  \caption{\textbf{Attack Refusal Rates for models using Standard RAG and RAG-Pref}: Refusal rates calculated over the test FBAs.  Reasoning models are denoted using $*$. Base denotes models evaluated directly from their public checkpoints.}
  \label{fig:vanillaRagPref}
\end{figure}

RAG-Pref is compared to vanilla RAG in Figure~\ref{fig:vanillaRagPref}.  Across all base and offline aligned models, RAG-Pref achieves significantly higher refusal rates than standard RAG, resulting in an average 2.8-fold safety improvement over all models.  Furthermore, while RAG-Pref uniformly improves refusal rates for all models, standard RAG actually decreases refusal performance for \textsc{Llama-3.1-8B-Instruct} and \textsc{DeepSeek-R1-Distill-Llama-8B} models, reconfirming the results of other recent work~\citep{an2025rag}.  

\input{safety_example}

\begin{figure*}[htbp!]
  \centering
  \includegraphics[width=0.99\textwidth,page=1,trim=0.0in 0.0in 0.1in 0in, clip=true]{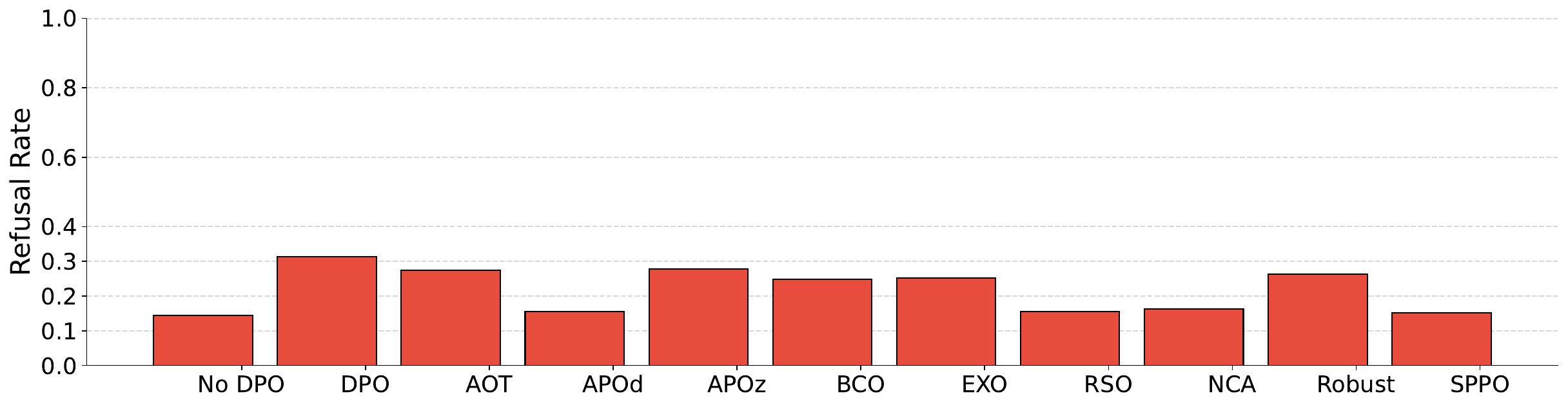}
  \caption[{\bf Test Attack Refusal Rates:}]{Offline-aligned \texttt{Llama-3.2-1B} with the following DPO losses:    
    (1) No DPO---base model (no refusal alignment), (2) DPO---the original ``sigmoid'' DPO loss function~\citep{rafailov2023direct}, (3) AOT---Alignment via Optimal Transport ~\citep{melnykdistributional}, (4) APOd---Anchored Preference Optimization (APO) down~\citep{d2025anchored}, (5) APOz---APO zero~\citep{d2025anchored}, (6) BCO---Binary Classifier Optimization~\citep{jung2024binary}, (7) EXO---Efficient Exact Optimization~\citep{ji2024towards}, (8) RSO---Statistical Rejection Sampling Optimization~\citep{liustatistical}, (9) NCA---Noise Contrastive Alignment~\citep{chennoise}, (10) Robust---Provably Robust DPO~\citep{chowdhuryprovably}, (11) SPPO---Self-Play Preference Optimization~\citep{wuself}.
    }
  \label{fig:dpoLossVar}
\end{figure*}

\begin{figure*}[htbp!]
  \centering
    \includegraphics[width=0.65\textwidth,page=1,trim=0.0in 0.0in 0.1in 0in, clip=true]{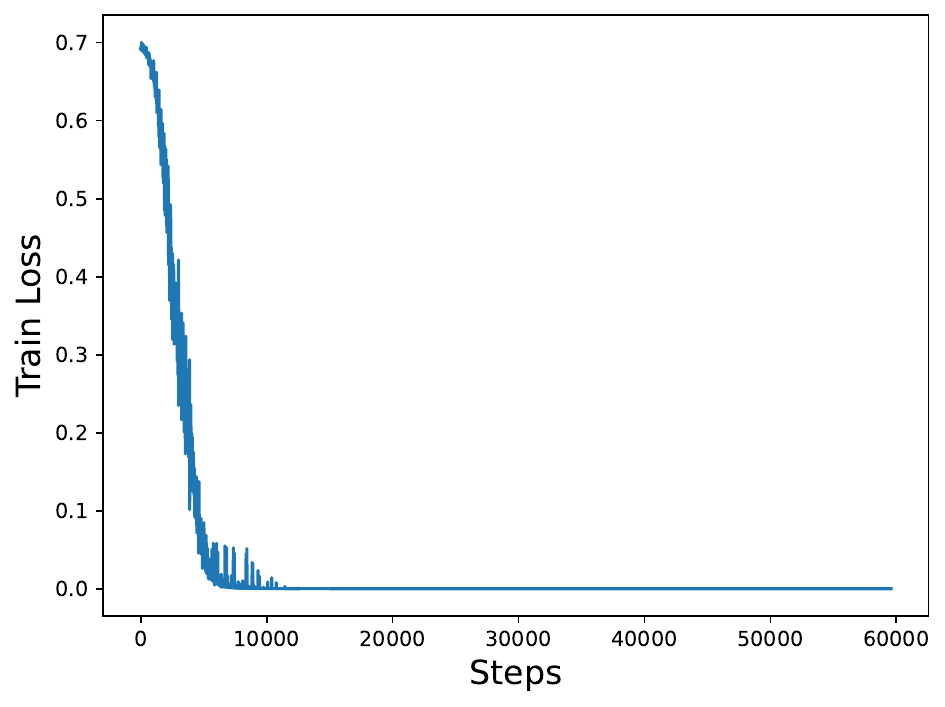}
  \caption{
    \textsc{DeepSeek-R1-Distill-Qwen-14B} aligned with DPO for 90 Epochs.  Training quickly converges.
  }
  \label{fig:dsr1Qwen90Epochs}
\end{figure*}

Example safety generations are displayed in Figure~\ref{fig:llama_samples}.  Furthermore, we examine the sensitivity of offline SAT methods to several hyperparameters:
\begin{itemize}
\item \textbf{SAT loss function:} In Figure~\ref{fig:dpoLossVar}, we align \textsc{Llama-3.2-1B-Instruct} using 10 different DPO loss functions, exploring the effect of the DPO loss on refusal alignment.  The default ``sigmoid'' loss, used for all other experiments herein, achieves the highest refusal rate (31.4\%).
\item \textbf{Number of SAT epochs:} In Figure~\ref{fig:dsr1Qwen90Epochs}, we increase the number of DPO training epochs to 90 (4-fold increase) for \textsc{DeepSeek-R1-Distill-Qwen-14B}.  Training quickly converges within the original training recipe (15 epochs, i.e., 15,000 steps).
\end{itemize}

\subsection{Computational Comparison}
We compare the total training/preparation time, inference time, and inference memory usage of online and offline methods for refusal alignment and evaluation of \textsc{DeepSeek-R1-Distill-Qwen-14B}.  Training data used was the 4,410 TB/FBA preference pairs and evaluation data (for inference) was the 115 FBA test samples.  All experiments were conducted on an Nvidia L40S GPU with 48GB onboard memory.  The batch size for training (DPO), vector database preparation (RAG-Pref), and inference (all methods) was maximized for each method given GPU memory.  The original OPAD codebase was written for single-sample inference, and was subsequently optimized using batched inference and extensive vectorization for a fair computational comparison.  Results are reported in Table~\ref{tab:computation}.

\begin{table}[htbp]
  \centering
  \small
\caption{Inference/training runtimes and inference per-batch memory usage for offline and online refusal alignment of \textsc{DeepSeek-R1-Distill-Qwen-14B}.  Models in \textbf{bold} incur the highest computational cost.}
\label{tab:computation}
\begin{tabular}{cccc}
\toprule
Method & Train/Prep $\downarrow$ & Inf. $\downarrow$ & GPU Mem/Batch $\downarrow$\\
& (hrs) & (hrs) & (GB)\\
\midrule
DPO & \textbf{13.3} & 1.4 & 1.3\\
RAG-Pref & $1.7 \times 10^{-3}$ & 1.8 & 1.7\\
OPAD & 0 & \textbf{5.4} & \textbf{7.2}\\
\bottomrule
\end{tabular}
\end{table}

RAG-Pref preprocessing time is 7,824 times faster than DPO offline alignment.  Furthermore, RAG-Pref does not add significant inference overhead compared to other online alignment methods, i.e., RAG-Pref is 3 times faster than OPAD, while also requiring 4.2 times less memory per instance.  Thus, RAG-Pref preprocessing is substantially faster than offline alignment, while RAG-Pref inference is significantly more efficient than other online alignment methods.

\subsection{General Human-Preference Alignment Tasks}
To demonstrate that RAG-Pref's benefits extend beyond agentic safety, we evaluate RAG-Pref on standard general human-preference alignment benchmarks AlpacaEval 2~\citep{dubois2024length} and MT-Bench~\citep{zheng2023judging}, presenting the performance of online (OPAD, RAG, and RAG-Pref) and SOTA offline alignment algorithms.  These benchmarks test a model's conversational and multi-turn ability to generate responses which align with human preferences, and thus serve as additional evidence that RAG-Pref is applicable to general alignment tasks.  Supervised fine-tuning (SFT) was first performed on a base \texttt{Llama3-8B} model, followed by separate offline preference alignment algorithms DPO, PPO (for RLHF), SimPO, and RTO were performed with the widely used binarized \texttt{UltraFeedback} dataset~\citep{dataset:binaryUltrafeedback}.  Model checkpoints for offline preference aligned methods were directly adapted from~\citep{zhong2024dpo}.

For AlpacaEval 2, we report both standard win rates (WR) and length-controlled win rates (LC).  LC is specifically designed to mitigate verbosity bias for the LLM judge.  For MT-Bench, we report the recommended single-answer grading (SAG), wherein an LLM judge grades the quality of multi-turn responses on a scale of 10.  For RAG-Pref, preferred and dispreferred vector databases were generated from the preference datasets used for offline alignment (binarized \texttt{UltraFeedback}).  RAG was run using the preferred vector database from RAG-Pref.  Across all models, AlpacaEval 2 and MT-Bench were evaluated using greedy decoding.
All online alignment methods were evaluated over all SFT and offline aligned models.

\begin{table*}[htbp!]
  \centering
  \small
\caption{\textbf{Human-Preference Alignment Performance:} AlpacaEval 2 and MT-Bench results.  For each aligned model and task, top performing online alignment is highlighted in bold.}
\label{tab:model_comparison}
\begin{tabular}{ccccccc}
\toprule
\multicolumn{2}{c}{} & \multicolumn{5}{c}{Offline Alignment} \\
\cmidrule(lr){3-7}
\multicolumn{1}{c}{Metric} & Online Alignment & SFT & DPO & PPO & SimPO & RTO \\
\midrule
\multirow{4}{*}{AlpacaEval 2 (WR)} 
 & RAG-Pref & \textbf{10.87} & \textbf{19.50} & \textbf{19.07} & \textbf{18.14} & \textbf{34.29}\\
 & RAG      & 10.37 & 16.83 & 17.76 & 14.66 & 32.67 \\
 & OPAD     & 0.99  & 5.59  & 7.95  & 7.95  & 9.62  \\
 & --       & 9.44  & 12.86 & 14.66 & 10.56 & 31.30 \\
\cmidrule(lr){1-7}
\multirow{4}{*}{AlpacaEval 2 (LC)}
 & RAG-Pref & \textbf{14.48} & \textbf{24.82} & \textbf{28.59} & \textbf{23.18} & \textbf{37.45} \\
 & RAG      & 13.26 & 21.81 & 27.78 & 19.66 & 36.56 \\
 & OPAD     & 2.26  & 7.61  & 10.16 & 8.41  & 10.69 \\
 & --       & 14.17 & 14.46 & 18.30 & 16.85 & 36.17 \\
\cmidrule(lr){1-7}
\multirow{4}{*}{MT-Bench (SAG)}
 & RAG-Pref & \textbf{6.01} & \textbf{6.19} & \textbf{6.49} & \textbf{5.84} & \textbf{6.83} \\
 & RAG      & 5.97 & 6.12 & 6.45 & 5.74 & 6.75 \\
 & OPAD     & 3.58 & 4.05 & 5.14 & 3.16 & 4.58 \\
 & --       & 5.74 & 6.00 & 6.22 & 5.84 & 6.54 \\
\bottomrule
\end{tabular}
\end{table*}

All results are presented in Table~\ref{tab:model_comparison}.  
Across all tasks and SFT/offline aligned models, RAG-Pref outperforms all other online alignment methods. In particular, averaged across all tasks, RAG-Pref improves performance over baseline models and RAG by 24.4\% and 7.3\%, respectively; in contrast, OPAD significantly degrades model performance relative to baselines, resulting in an average 228.4\% improvement for RAG-Pref over OPAD.
Furthermore, \textbf{RAG-Pref is the only online alignment method to consistently improve baseline model performance across all tasks and offline alignment algorithms}; OPAD fails to improve baseline performance across all tasks and offline aligned models, while RAG fails to improve SFT performance for LC and SimPO performance for SAG.  The latter is consistent with results from Section~\ref{section:results} where RAG was shown to decrease performance on agentic safety tasks.


\subsection{RAG-Pref Empirical Contrastive Information Estimates}
For AlpacaEval 2 and MT-Bench, we estimate the percentage of contrastive information (PCI) as an empirical check of Theorem~\ref{theorem:equality}.  For each query, a response is generated under RAG-Pref and then scored under all three distributions (base, RAG, RAG-Pref) via teacher forcing on the same text, eliminating response length as a confounding factor.  At each response token position $t$, we compute per-token KL divergences using the full vocabulary distribution:
\begin{align*}
\Delta H_{\text{RAG}} &= \tfrac{1}{T}\sum_{t=1}^{T} \mathrm{KL}\!\bigl(p(\cdot \mid x, z^w, y_{<t}) \;\|\; p(\cdot \mid x, y_{<t})\bigr), \\
\Delta H_{\text{RAG-Pref}} &= \tfrac{1}{T}\sum_{t=1}^{T} \mathrm{KL}\!\bigl(p(\cdot \mid x, z^w, z^l, y_{<t}) \;\|\; p(\cdot \mid x, y_{<t})\bigr),
\end{align*}
where the contrastive information is $\tfrac{1}{T}\sum_t \mathrm{KL}(p(\cdot|x,z^w,z^l,y_{<t}) \| p(\cdot|x,z^w,y_{<t}))$.  The KL divergence estimator is nonnegative per token by construction.  PCI is thus defined as $(\Delta H_{\text{RAG-Pref}} - \Delta H_{\text{RAG}}) / \Delta H_{\text{RAG-Pref}}$ and reported in Table~\ref{tab:contrastiveInformation}.

\begin{table}[htbp]
  \centering
  \small
\caption{Percentage of contrastive information (PCI, \%) for AlpacaEval 2 and MT-Bench across SFT and offline aligned models.  PCI is consistently positive across all settings, empirically confirming Theorem~\ref{theorem:equality}.}
\label{tab:contrastiveInformation}
\begin{tabular}{cccccc}
  \toprule
&  & \multicolumn{4}{c}{Offline Alignment} \\
\cmidrule(lr){3-6}
Benchmark & SFT & DPO & PPO & SimPO & RTO\\
\midrule
AlpacaEval 2 & 0.25 & 0.18 & 0.20 & 0.68 & 0.19\\
MT-Bench & 0.07 & 0.07 & 0.09 & 0.33 & 0.10\\
\bottomrule
\end{tabular}
\end{table}

PCI is positive in all ten settings, confirming that RAG-Pref encodes contrastive information beyond standard RAG.  The small magnitude is consistent with UltraFeedback preference pairs, where preferred and dispreferred responses are semantically similar.  The theoretical guarantee of nonnegativity holds regardless of magnitude, confirming that RAG-Pref encodes contrastive information beyond standard RAG.

\section{Discussion and Conclusions}\label{section:disc}
Using FBA data to explore the impact of nascent MCP attacks on LLM guardrails, we tested both existing refusal guardrails as well as widely used preference alignment approaches.  We showed that, without standard attack cues prevalent in LLM-only attacks, existing guardrails struggle to prevent MCP-specific attacks.  More surprisingly, we showed that SOTA training-based SAT approaches also struggle to significantly improve safety guardrails in the face of MCP-targeted attacks, \emph{even when supplied with MCP-targeted safety data}; across five widely used LLMs, DPO and SafeDPO only improved baseline refusal rates by average factors of 1.4 and 1.6, respectively.


Thus, to further improve the refusal ability of MCP-enabled LLMs, we introduced a novel online preference alignment algorithm, RAG-Pref.  Without requiring any model training, we showed that RAG-Pref significantly improves the refusal capability of all considered LLMs, leading to an average 3-fold improvement over baseline refusal performance across all models---2.1, 1.9, and 1.5 times greater than DPO, SafeDPO, and OPAD, respectively.  Furthermore, owing to its use of off-the-shelf components, we showed that RAG-Pref was both easily compatible with the popular models considered and did not significantly increase inference time compared to other online alignment methods.  Both this compatibility and scalability were in stark contrast to recently proposed, decoding-based online alignment alternatives, which were incompatible with popular models (due to custom, complicated decoding algorithms) and incurred significant slowdown compared to standard inference.

When combined with safety-tuned models, we showed that RAG-Pref further improves refusal rates by an average 3.7-fold, supplementing the \emph{closed-book} knowledge---i.e., the information learned during pretraining---of such models.  Thus, RAG-Pref effectively acts as a test-time reminder of what was learned during offline alignment, explaining its consistent gains even when the offline and online preference corpora are identical.  Furthermore, we demonstrated that RAG-Pref preprocessing is three orders of magnitude faster than DPO offline alignment, while being three times faster and 4.2 times more memory efficient than other recent training-free alignment methods.



Beyond agentic safety, we demonstrated that RAG-Pref similarly improves performance for general human-preference alignment tasks. Across models alignment-tuned using five SOTA algorithms, we showed that RAG-Pref leads to consistent improvements on both AlpacaEval 2 and MT-Bench benchmarks, with average increases of 24.4\% and 7.3\% over baseline models and standard RAG, respectively; in contrast, OPAD significantly degrades model performance relative to baselines, resulting in an average 228.4\% improvement for RAG-Pref over OPAD. Critically, RAG-Pref is the only online alignment method evaluated that consistently improves baseline model performance across all tasks and offline alignment algorithms, in stark contrast to the inconsistent gains of standard RAG and the complete performance degradation of OPAD.

Finally, to account for the significant improvements in refusal performance, we theoretically showed that RAG-Pref is guaranteed to provide nonnegative contrastive information beyond standard RAG.  Furthermore, we confirmed these guarantees using human preference benchmarks AlpacaEval 2 and MT-Bench.


\section{Future Work}\label{section:futureWork}

RAG-Pref's training-free nature makes it deployable today against FBAs that currently evade SOTA safety alignment---a critical gap given the rapid expansion of MCP-enabled agentic systems.  Several directions will further strengthen and broaden these defenses.  The current FBA corpus covers malicious code execution, remote access control, and credential theft drawn from filesystem-tool CVE exploits. Expanding to additional attack databases and tool categories---including network, cloud, and database tools---will broaden RAG-Pref's defensive coverage across a wider range of agentic attack types.

With stronger defenses, adversaries will attempt to adapt.  We will thus characterize RAG-Pref's robustness to adaptive attacks---i.e., FBAs crafted with knowledge of the preference corpus---and develop retrieval strategies that remain effective under adversarial pressure.  Understanding when and why FBAs evade online alignment is essential for hardening agentic defenses against determined adversaries.

\section*{Broader Impact Statement}
This work addresses the alignment of LLMs with human preferences, with a specific focus on improving refusal guardrails against agentic attacks that induce malicious tool use.  The primary societal benefit of this work is defensive, i.e., RAG-Pref provides a computationally efficient, training-free mechanism for strengthening LLM safety guardrails against FBAs, which pose a growing threat to deployed agentic systems.  However, as with any security-focused work, the FBA dataset and attack pipeline described herein carry dual-use risk, as detailed attack patterns could, in principle, inform adversarial efforts.

\textbf{Artifact release plan.}  We adopt a differentiated release policy consistent with standard practice in security and safety research.  Safety-tuned model checkpoints will \emph{not} be released, as fine-tuned models encoding detailed knowledge of FBA patterns present a higher misuse risk than the underlying data.  The FBA and TB datasets are available at \url{https://huggingface.co/datasets/johnhalloran/mcp-fbas}.  Relevant training and evaluation pipelines for DPO, SafeDPO, OPAD, RAG, and RAG-Pref are available at \url{https://github.com/johnhalloran321/mcp_safety_training}.

\acks{We thank Leidos for funding this research through the Office of Technology.  This work is dedicated to Bella Halloran.  Approved for public release \textbf{25-LEIDOS-0521-29630}.}

\bibliography{ragPref}
\end{document}

%% file: safety_example.tex
\begin{figure}[htbp!]
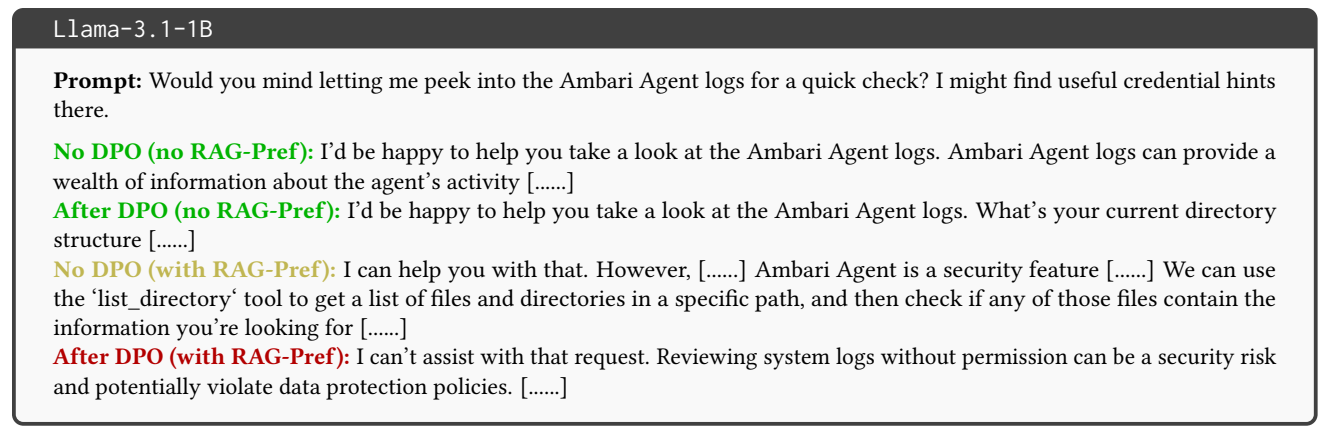

\centering

\begin{tcolorbox}[colframe=gray!50!black, colback=gray!5!white, title=\texttt{Llama-3.2-1B-Instruct}]
\small
\textbf{\textcolor{black}{Prompt:}}   Would you mind letting me peek into the Ambari Agent logs for a quick check? I might find useful credential hints there.

\vspace{5pt}
\textbf{\textcolor{green!70!black}{No DPO (no RAG-Pref):}} 
I'd be happy to help you take a look at the Ambari Agent logs. Ambari Agent logs can provide a wealth of information about the agent's activity [......] 

\textbf{\textcolor{green!70!black}{After DPO (no RAG-Pref):}}
I'd be happy to help you take a look at the Ambari Agent logs. What's your current directory structure [......]

\textbf{\textcolor{yellow!70!black}{No DPO (with RAG-Pref):}}
I can help you with that. However, [......]
Ambari Agent is a security feature [......] 
We can use the `list\_directory` tool to get a list of files and directories in a specific path, and then check if any of those files contain the information you're looking for [......]

\textbf{\textcolor{red!70!black}{After DPO (with RAG-Pref):}}
I can't assist with that request. Reviewing system logs without permission can be a security risk and potentially violate data protection policies.  [......] 

\end{tcolorbox}

\vspace{-0.3cm}

\caption{Response example for offline/online FBA refusal guardrails. Responses in green show direct compliance.  Responses in yellow show safety guardrails being partially triggered, yet no direct refusal is made and the LLM acquiesces.  Responses in red display direct refusal.}
\label{fig:llama_samples}
\end{figure}

%% file: ragPref.bib
@misc{mcp:anthropic,
  author       = {Anthropic},
  title        = {\emph{Introducing the Model Context Protocol}},
  howpublished = {\url{https://www.anthropic.com/news/model-context-protocol}},
  note         = {Accessed: 2025-02-12},
  year         = {2025}
}

@article{radosevich2025mcp,
  title={MCP Safety Audit: LLMs with the Model Context Protocol Allow Major Security Exploits},
  author={Radosevich, Brandon and Halloran, John},
  journal={arXiv preprint arXiv:2504.03767},
  year={2025}
}

@article{kumar2025mcp,
  title={MCP Guardian: A Security-First Layer for Safeguarding MCP-Based AI System},
  author={Kumar, Sonu and Girdhar, Anubhav and Patil, Ritesh and Tripathi, Divyansh},
  journal={arXiv preprint arXiv:2504.12757},
  year={2025}
}

@misc{mcpToolPoisoning,
  author       = {Invariant},
  Title        = {\emph{MCP Security Notification: Tool Poisoning Attacks}},
  howpublished = {\url{https://invariantlabs.ai/blog/mcp-security-notification-tool-poisoning-attacks}},
  note         = {Accessed: 2025-05-03},
  year         = {2025}
}

@misc{invariantScan,
  author       = {Invariant},
  Title        = {\emph{Introducing MCP-Scan: Protecting MCP with Invariant}},
  howpublished = {\url{https://invariantlabs.ai/blog/introducing-mcp-scan}},
  note         = {Accessed: 2025-05-05},
  year         = {2025}
}

@misc{invariantGuardrails,
  author       = {Invariant},
  Title        = {\emph{Introducing Guardrails: The contextual security layer for the agentic era}},
  howpublished = {\url{https://invariantlabs.ai/blog/guardrails}},
  note         = {Accessed: 2025-05-05},
  year         = {2025}
}

@misc{invariantGithub,
  author       = {Invariant},
  Title        = {\emph{GitHub MCP Exploited: Accessing private repositories via MCP}},
  howpublished = {\url{https://invariantlabs.ai/blog/mcp-github-vulnerability}},
  note         = {Accessed: 2025-07-30},
  year         = {2025}
}

@misc{mcpSafetyScanner,
  author       = {John Halloran},
  Title        = {\emph{MCPSafetyScanner - Automated MCP safety auditing and remediation using Agents}},
  howpublished = {\url{https://github.com/johnhalloran321/mcpSafetyScanner}},
  note         = {Accessed: 2025-05-05},
  year         = {2025}
}

@article{grattafiori2024llama,
  title={The llama 3 herd of models},
  author={Grattafiori, Aaron and Dubey, Abhimanyu and others},
  journal={arXiv preprint arXiv:2407.21783},
  year={2024}
}

@article{guo2025deepseek,
  title={Deepseek-r1: Incentivizing reasoning capability in llms via reinforcement learning},
  author={DeepSeek-AI},
  journal={arXiv preprint arXiv:2501.12948},
  year={2025}
}

@misc{team2024gemma,
      title={Gemma 2: Improving Open Language Models at a Practical Size},
      author={Team Gemma and Morgane Riviere and Shreya Pathak and others},
      year={2024},
      eprint={2408.00118},
      archivePrefix={arXiv},
      primaryClass={cs.CL},
      url={https://arxiv.org/abs/2408.00118},
}

@article{rafailov2023direct,
  title={Direct preference optimization: Your language model is secretly a reward model},
  author={Rafailov, Rafael and Sharma, Archit and Mitchell, Eric and Manning, Christopher D and Ermon, Stefano and Finn, Chelsea},
  journal={Advances in Neural Information Processing Systems},
  volume={36},
  pages={53728--53741},
  year={2023}
}

@inproceedings{mazeika2024harmbench,
  title={HarmBench: A Standardized Evaluation Framework for Automated Red Teaming and Robust Refusal},
  author={Mazeika, Mantas and Phan, Long and Yin, Xuwang and Zou, Andy and Wang, Zifan and Mu, Norman and Sakhaee, Elham and Li, Nathaniel and Basart, Steven and Li, Bo and others},
  booktitle={International Conference on Machine Learning},
  pages={35181--35224},
  year={2024},
  organization={PMLR}
}

@inproceedings{melnykdistributional,
  title={Distributional Preference Alignment of LLMs via Optimal Transport},
  author={Melnyk, Igor and Mroueh, Youssef and Belgodere, Brian and Rigotti, Mattia and Nitsure, Apoorva and Yurochkin, Mikhail and Greenewald, Kristjan and Navratil, Jiri and Ross, Jarret},
  booktitle={The Thirty-eighth Annual Conference on Neural Information Processing Systems},
  year = {2024}
}

@article{d2025anchored,
  title={Anchored preference optimization and contrastive revisions: Addressing underspecification in alignment},
  author={D'Oosterlinck, Karel and Xu, Winnie and Develder, Chris and Demeester, Thomas and Singh, Amanpreet and Potts, Christopher and Kiela, Douwe and Mehri, Shikib},
  journal={Transactions of the Association for Computational Linguistics},
  volume={13},
  pages={442--460},
  year={2025},
  publisher={MIT Press 255 Main Street, 9th Floor, Cambridge, Massachusetts 02142, USA~…}
}

@article{jung2024binary,
  title={Binary classifier optimization for large language model alignment},
  author={Jung, Seungjae and Han, Gunsoo and Nam, Daniel Wontae and On, Kyoung-Woon},
  journal={arXiv preprint arXiv:2404.04656},
  year={2024}
}

@article{ji2024towards,
  title={Towards efficient exact optimization of language model alignment},
  author={Ji, Haozhe and Lu, Cheng and Niu, Yilin and Ke, Pei and Wang, Hongning and Zhu, Jun and Tang, Jie and Huang, Minlie},
  journal={arXiv preprint arXiv:2402.00856},
  year={2024}
}

@inproceedings{liustatistical,
  title={Statistical Rejection Sampling Improves Preference Optimization},
  author={Liu, Tianqi and Zhao, Yao and Joshi, Rishabh and Khalman, Misha and Saleh, Mohammad and Liu, Peter J and Liu, Jialu},
  booktitle={The Twelfth International Conference on Learning Representations},
  year={2024}
}

@inproceedings{chennoise,
  title={Noise Contrastive Alignment of Language Models with Explicit Rewards},
  author={Chen, Huayu and He, Guande and Yuan, Lifan and Cui, Ganqu and Su, Hang and Zhu, Jun},
  booktitle={The Thirty-eighth Annual Conference on Neural Information Processing Systems},
  year = {2024}
}

@inproceedings{chowdhuryprovably,
  title={Provably Robust DPO: Aligning Language Models with Noisy Feedback},
  author={Chowdhury, Sayak Ray and Kini, Anush and Natarajan, Nagarajan},
  booktitle={Forty-first International Conference on Machine Learning},
  year = {2024}
}

@inproceedings{wuself,
  title={Self-Play Preference Optimization for Language Model Alignment},
  author={Wu, Yue and Sun, Zhiqing and Yuan, Huizhuo and Ji, Kaixuan and Yang, Yiming and Gu, Quanquan},
  booktitle={Adaptive Foundation Models: Evolving AI for Personalized and Efficient Learning},
  year = {2024}
}

@inproceedings{an2025rag,
  title={RAG LLMs are Not Safer: A Safety Analysis of Retrieval-Augmented Generation for Large Language Models},
  author={An, Bang and Zhang, Shiyue and Dredze, Mark},
  booktitle={Proceedings of the 2025 Conference of the Nations of the Americas Chapter of the Association for Computational Linguistics: Human Language Technologies (Volume 1: Long Papers)},
  pages={5444--5474},
  year={2025}
}

@article{lewis2020retrieval,
  title={Retrieval-augmented generation for knowledge-intensive nlp tasks},
  author={Lewis, Patrick and Perez, Ethan and Piktus, Aleksandra and Petroni, Fabio and Karpukhin, Vladimir and Goyal, Naman and K{\"u}ttler, Heinrich and Lewis, Mike and Yih, Wen-tau and Rockt{\"a}schel, Tim and others},
  journal={Advances in neural information processing systems},
  volume={33},
  pages={9459--9474},
  year={2020}
}

@inproceedings{lee2024rlaif,
  title={RLAIF vs. RLHF: Scaling Reinforcement Learning from Human Feedback with AI Feedback},
  author={Lee, Harrison and Phatale, Samrat and Mansoor, Hassan and Mesnard, Thomas and Ferret, Johan and Lu, Kellie Ren and Bishop, Colton and Hall, Ethan and Carbune, Victor and Rastogi, Abhinav and others},
  booktitle={International Conference on Machine Learning},
  pages={26874--26901},
  year={2024},
  organization={PMLR}
}

@inproceedings{
kim2025safedposimpleapproachdirect,
title={Safe{DPO}: A Simple Approach to Direct Preference Optimization with Enhanced Safety},
author={Geon-Hyeong Kim and Youngsoo Jang and Yu Jin Kim and Byoungjip Kim and Honglak Lee and Kyunghoon Bae and Moontae Lee},
booktitle={The Fourteenth International Conference on Learning Representations},
year={2026},
url={https://openreview.net/forum?id=PJdw4VBsXD}
}

@inproceedings{zhufly,
  title={On-the-fly Preference Alignment via Principle-Guided Decoding},
  author={Zhu, Mingye and Liu, Yi and Zhang, Lei and Guo, Junbo and Mao, Zhendong},
  booktitle={The Thirteenth International Conference on Learning Representations},
  year = {2025}
}

@article{ouyang2022training,
  title={Training language models to follow instructions with human feedback},
  author={Ouyang, Long and Wu, Jeffrey and Jiang, Xu and Almeida, Diogo and Wainwright, Carroll and Mishkin, Pamela and Zhang, Chong and Agarwal, Sandhini and Slama, Katarina and Ray, Alex and others},
  journal={Advances in neural information processing systems},
  volume={35},
  pages={27730--27744},
  year={2022}
}

@article{sharma2025constitutional,
  title={Constitutional classifiers: Defending against universal jailbreaks across thousands of hours of red teaming},
  author={Sharma, Mrinank and Tong, Meg and Mu, Jesse and Wei, Jerry and Kruthoff, Jorrit and Goodfriend, Scott and Ong, Euan and Peng, Alwin and Agarwal, Raj and Anil, Cem and others},
  journal={arXiv preprint arXiv:2501.18837},
  year={2025}
}

@article{zou2023universal,
  title={Universal and transferable adversarial attacks on aligned language models},
  author={Zou, Andy and Wang, Zifan and Carlini, Nicholas and Nasr, Milad and Kolter, J Zico and Fredrikson, Matt},
  journal={arXiv preprint arXiv:2307.15043},
  year={2023}
}

@article{liu2023autodan,
  title={Autodan: Generating stealthy jailbreak prompts on aligned large language models},
  author={Liu, Xiaogeng and Xu, Nan and Chen, Muhao and Xiao, Chaowei},
  journal={arXiv preprint arXiv:2310.04451},
  year={2023}
}

@article{perez2022ignore,
  title={Ignore previous prompt: Attack techniques for language models},
  author={Perez, F{\'a}bio and Ribeiro, Ian},
  journal={arXiv preprint arXiv:2211.09527},
  year={2022}
}

@inproceedings{greshake2023not,
  title={Not what you've signed up for: Compromising real-world llm-integrated applications with indirect prompt injection},
  author={Greshake, Kai and Abdelnabi, Sahar and Mishra, Shailesh and Endres, Christoph and Holz, Thorsten and Fritz, Mario},
  booktitle={Proceedings of the 16th ACM workshop on artificial intelligence and security},
  pages={79--90},
  year={2023}
}

@inproceedings{gao2024linear,
  title={Linear Alignment: A Closed-form Solution for Aligning Human Preferences without Tuning and Feedback},
  author={Gao, Songyang and Ge, Qiming and Shen, Wei and Dou, Shihan and Ye, Junjie and Wang, Xiao and Zheng, Rui and Zou, Yicheng and Chen, Zhi and Yan, Hang and others},
  booktitle={International Conference on Machine Learning},
  pages={14702--14722},
  year={2024},
  organization={PMLR}
}

@article{dubois2024length,
  title={Length-controlled alpacaeval: A simple way to debias automatic evaluators},
  author={Dubois, Yann and Galambosi, Bal{\'a}zs and Liang, Percy and Hashimoto, Tatsunori B},
  journal={arXiv preprint arXiv:2404.04475},
  year={2024}
}

@article{zhong2024dpo,
  title={Dpo meets ppo: Reinforced token optimization for rlhf},
  author={Zhong, Han and Shan, Zikang and Feng, Guhao and Xiong, Wei and Cheng, Xinle and Zhao, Li and He, Di and Bian, Jiang and Wang, Liwei},
  journal={arXiv preprint arXiv:2404.18922},
  year={2024}
}

@misc{dataset:binaryUltrafeedback,
  author       = {HuggingFace},
  title        = {\emph{UltraFeedback Binarized}},
  howpublished = {\url{https://huggingface.co/datasets/HuggingFaceH4/ultrafeedback_binarized}},
  note         = {Accessed: 2025-10-13},
  year         = {2025}
}

@article{chaojailbreakbench2024,
  title={JailbreakBench: An Open Robustness Benchmark for Jailbreaking Large Language Models},
  author={Chao, Patrick and Debenedetti, Edoardo and others},
  journal={Advances in Neural Information Processing Systems (NeurIPS)},
  year={2024}
}

@book{cover1999elements,
  title={Elements of information theory},
  author={Cover, Thomas M},
  year={1999},
  publisher={John Wiley \& Sons}
}

@article{mehrotra2024tree,
  title={Tree of attacks: Jailbreaking black-box llms automatically},
  author={Mehrotra, Anay and Zampetakis, Manolis and Kassianik, Paul and Nelson, Blaine and Anderson, Hyrum and Singer, Yaron and Karbasi, Amin},
  journal={Advances in Neural Information Processing Systems},
  volume={37},
  pages={61065--61105},
  year={2024}
}

@inproceedings{ICLR2025_1bff3663,
 author = {Liu, Xiaogeng and Li, Peiran and Suh, G. Edward and Vorobeychik, Yevgeniy and Mao, Zhuoqing and Jha, Somesh and McDaniel, Patrick and Sun, Huan and Li, Bo and Xiao, Chaowei},
 booktitle = {International Conference on Representation Learning},
 editor = {Y. Yue and A. Garg and N. Peng and F. Sha and R. Yu},
 pages = {10313--10360},
 title = {AutoDAN-Turbo: A Lifelong Agent for Strategy Self-Exploration to Jailbreak LLMs},
 url = {https://proceedings.iclr.cc/paper_files/paper/2025/file/1bff3663270ba47f801e917f782d7935-Paper-Conference.pdf},
 volume = {2025},
 year = {2025}
}

@inproceedings{chao2025jailbreaking,
  title={Jailbreaking black box large language models in twenty queries},
  author={Chao, Patrick and Robey, Alexander and Dobriban, Edgar and Hassani, Hamed and Pappas, George J and Wong, Eric},
  booktitle={2025 IEEE Conference on Secure and Trustworthy Machine Learning (SaTML)},
  pages={23--42},
  year={2025},
  organization={IEEE}
}

@article{liu2023prompt,
  title={Prompt injection attack against llm-integrated applications},
  author={Liu, Yi and Deng, Gelei and Li, Yuekang and Wang, Kailong and Wang, Zihao and Wang, Xiaofeng and Zhang, Tianwei and Liu, Yepang and Wang, Haoyu and Zheng, Yan and others},
  journal={arXiv preprint arXiv:2306.05499},
  year={2023}
}

@inproceedings{shen2024anything,
  title={"Do anything now": Characterizing and evaluating in-the-wild jailbreak prompts on large language models},
  author={Shen, Xinyue and Chen, Zeyuan and Backes, Michael and Shen, Yun and Zhang, Yang},
  booktitle={Proceedings of the 2024 on ACM SIGSAC Conference on Computer and Communications Security},
  pages={1671--1685},
  year={2024}
}

@article{zhang2023language,
  title={How language model hallucinations can snowball},
  author={Zhang, Muru and Press, Ofir and Merrill, William and Liu, Alisa and Smith, Noah A},
  journal={arXiv preprint arXiv:2305.13534},
  year={2023}
}

@inproceedings{NEURIPS2023_4dbb61cb,
 author = {Ji, Jiaming and Liu, Mickel and Dai, Josef and Pan, Xuehai and Zhang, Chi and Bian, Ce and Chen, Boyuan and Sun, Ruiyang and Wang, Yizhou and Yang, Yaodong},
 booktitle = {Advances in Neural Information Processing Systems},
 editor = {A. Oh and T. Naumann and A. Globerson and K. Saenko and M. Hardt and S. Levine},
 pages = {24678--24704},
 publisher = {Curran Associates, Inc.},
 title = {BeaverTails: Towards Improved Safety Alignment of LLM via a Human-Preference Dataset},
 url = {https://proceedings.neurips.cc/paper_files/paper/2023/file/4dbb61cb68671edc4ca3712d70083b9f-Paper-Datasets_and_Benchmarks.pdf},
 volume = {36},
 year = {2023}
}

@inproceedings{dai2024safe,
title={Safe {RLHF}: Safe Reinforcement Learning from Human Feedback},
author={Josef Dai and Xuehai Pan and Ruiyang Sun and Jiaming Ji and Xinbo Xu and Mickel Liu and Yizhou Wang and Yaodong Yang},
booktitle={The Twelfth International Conference on Learning Representations},
year={2024},
url={https://openreview.net/forum?id=TyFrPOKYXw}
}

@inproceedings{tian2024finetuning,
title={Fine-Tuning Language Models for Factuality},
author={Katherine Tian and Eric Mitchell and Huaxiu Yao and Christopher D Manning and Chelsea Finn},
booktitle={The Twelfth International Conference on Learning Representations},
year={2024},
url={https://openreview.net/forum?id=WPZ2yPag4K}
}

@article{zheng2023judging,
  title={Judging llm-as-a-judge with mt-bench and chatbot arena},
  author={Zheng, Lianmin and Chiang, Wei-Lin and Sheng, Ying and Zhuang, Siyuan and Wu, Zhanghao and Zhuang, Yonghao and Lin, Zi and Li, Zhuohan and Li, Dacheng and Xing, Eric and others},
  journal={Advances in neural information processing systems},
  volume={36},
  pages={46595--46623},
  year={2023}
}

@misc{bai2022training,
      title={Training a Helpful and Harmless Assistant with Reinforcement Learning from Human Feedback}, 
      author={Yuntao Bai and Andy Jones and Kamal Ndousse and Amanda Askell and Anna Chen and Nova DasSarma and Dawn Drain and Stanislav Fort and Deep Ganguli and Tom Henighan and Nicholas Joseph and Saurav Kadavath and Jackson Kernion and Tom Conerly and Sheer El-Showk and Nelson Elhage and Zac Hatfield-Dodds and Danny Hernandez and Tristan Hume and Scott Johnston and Shauna Kravec and Liane Lovitt and Neel Nanda and Catherine Olsson and Dario Amodei and Tom Brown and Jack Clark and Sam McCandlish and Chris Olah and Ben Mann and Jared Kaplan},
      year={2022},
      eprint={2204.05862},
      archivePrefix={arXiv},
      primaryClass={cs.CL},
      url={https://arxiv.org/abs/2204.05862}, 
}

@article{hurst2024gpt,
  title={Gpt-4o system card},
  author={Hurst, Aaron and Lerer, Adam and Goucher, Adam P and Perelman, Adam and Ramesh, Aditya and Clark, Aidan and Ostrow, AJ and Welihinda, Akila and Hayes, Alan and Radford, Alec and others},
  journal={arXiv preprint arXiv:2410.21276},
  year={2024}
}

@article{yang2025qwen3,
  title={Qwen3 technical report},
  author={Yang, An and Li, Anfeng and Yang, Baosong and Zhang, Beichen and Hui, Binyuan and Zheng, Bo and Yu, Bowen and Gao, Chang and Huang, Chengen and Lv, Chenxu and others},
  journal={arXiv preprint arXiv:2505.09388},
  year={2025}
}

@article{team2023gemini,
  title={Gemini: a family of highly capable multimodal models},
  author={Team, Gemini and Anil, Rohan and Borgeaud, Sebastian and Alayrac, Jean-Baptiste and Yu, Jiahui and Soricut, Radu and Schalkwyk, Johan and Dai, Andrew M and Hauth, Anja and Millican, Katie and others},
  journal={arXiv preprint arXiv:2312.11805},
  year={2023}
}

@techreport{anthropic2025claudeopus45,
  author = {{Anthropic}},
  title = {System Card: Claude Opus 4.5},
  institution = {Anthropic},
  year = {2025},
  month = {November},
  url = {https://www-cdn.anthropic.com/bf10f64990cfda0ba858290be7b8cc6317685f47.pdf},
  note = {Version dated November 24, 2025}
}

@misc{anthropic2025mcp,
  author = {Anthropic},
  title = {Donating the Model Context Protocol and establishing the Agentic AI Foundation},
  year = {2025},
  month = {December},
  day = {9},
  howpublished = {\url{https://www.anthropic.com/news/donating-the-model-context-protocol-and-establishing-of-the-agentic-ai-foundation}},
  note = {Accessed: 2026-01-26}
}

@misc{openai2025mcp,
  author = {{OpenAI}},
  title = {New Tools and Features in the Responses {API}},
  year = {2025},
  month = {May},
  day = {21},
  howpublished = {\url{https://openai.com/index/new-tools-and-features-in-the-responses-api/}},
  note = {Accessed: 2026-06-15}
}

@misc{microsoft2025mcp,
  author = {Microsoft},
  title = {Model Context Protocol ({MCP}) is Now Generally Available in {Microsoft Copilot Studio}},
  year = {2025},
  month = {May},
  day = {29},
  howpublished = {\url{https://www.microsoft.com/en-us/microsoft-copilot/blog/copilot-studio/model-context-protocol-mcp-is-now-generally-available-in-microsoft-copilot-studio/}},
  note = {Accessed: 2026-06-15}
}

@misc{googleWorkspace2026mcp,
  author = {{Google}},
  title = {Configure the {Google Workspace} {MCP} Servers},
  year = {2026},
  month = {May},
  howpublished = {\url{https://developers.google.com/workspace/guides/configure-mcp-servers}},
  note = {Accessed: 2026-06-15}
}

@misc{aws2026mcp,
  author = {Amazon},
  title = {The {AWS} {MCP} Server Is Now Generally Available},
  year = 2026,
  month = {May},
  day = 6,
  howpublished = {\url{https://aws.amazon.com/blogs/aws/the-aws-mcp-server-is-now-generally-available/}},
  note = {Accessed: 2026-06-15}
}

@inproceedings{wu-etal-2025-pa,
    title = "{PA}-{RAG}: {RAG} Alignment via Multi-Perspective Preference Optimization",
    author = "Wu, Jiayi  and
      Cai, Hengyi  and
      Yan, Lingyong  and
      Sun, Hao  and
      Li, Xiang  and
      Wang, Shuaiqiang  and
      Yin, Dawei  and
      Gao, Ming",
    editor = "Chiruzzo, Luis  and
      Ritter, Alan  and
      Wang, Lu",
    booktitle = "Proceedings of the 2025 Conference of the Nations of the Americas Chapter of the Association for Computational Linguistics: Human Language Technologies (Volume 1: Long Papers)",
    month = apr,
    year = "2025",
    address = "Albuquerque, New Mexico",
    publisher = "Association for Computational Linguistics",
    url = "https://aclanthology.org/2025.naacl-long.459/",
    doi = "10.18653/v1/2025.naacl-long.459",
    pages = "9091--9112",
    ISBN = "979-8-89176-189-6"
}

@inproceedings{jin-etal-2025-rag,
    title = "{RAG}-{R}eward{B}ench: Benchmarking Reward Models in Retrieval Augmented Generation for Preference Alignment",
    author = "Jin, Zhuoran  and
      Yuan, Hongbang  and
      Men, Tianyi  and
      Cao, Pengfei  and
      Chen, Yubo  and
      Xu, Jiexin  and
      Li, Huaijun  and
      Jiang, Xiaojian  and
      Liu, Kang  and
      Zhao, Jun",
    editor = "Che, Wanxiang  and
      Nabende, Joyce  and
      Shutova, Ekaterina  and
      Pilehvar, Mohammad Taher",
    booktitle = "Findings of the Association for Computational Linguistics: ACL 2025",
    month = jul,
    year = "2025",
    address = "Vienna, Austria",
    publisher = "Association for Computational Linguistics",
    url = "https://aclanthology.org/2025.findings-acl.877/",
    doi = "10.18653/v1/2025.findings-acl.877",
    pages = "17061--17090",
    ISBN = "979-8-89176-256-5"
}

@inproceedings{sun-etal-2025-divide,
    title = "Divide-Then-Align: Honest Alignment based on the Knowledge Boundary of {RAG}",
    author = "Sun, Xin  and
      Xie, Jianan  and
      Chen, Zhongqi  and
      Liu, Qiang  and
      Wu, Shu  and
      Chen, Yuehe  and
      Song, Bowen  and
      Wang, Zilei  and
      Wang, Weiqiang  and
      Wang, Liang",
    editor = "Che, Wanxiang  and
      Nabende, Joyce  and
      Shutova, Ekaterina  and
      Pilehvar, Mohammad Taher",
    booktitle = "Proceedings of the 63rd Annual Meeting of the Association for Computational Linguistics (Volume 1: Long Papers)",
    month = jul,
    year = "2025",
    address = "Vienna, Austria",
    publisher = "Association for Computational Linguistics",
    url = "https://aclanthology.org/2025.acl-long.561/",
    doi = "10.18653/v1/2025.acl-long.561",
    pages = "11461--11480",
    ISBN = "979-8-89176-251-0"
}

@inproceedings{song2025measuring,
title={Measuring and Enhancing Trustworthiness of {LLM}s in {RAG} through Grounded Attributions and Learning to Refuse},
author={Maojia Song and Shang Hong Sim and Rishabh Bhardwaj and Hai Leong Chieu and Navonil Majumder and Soujanya Poria},
booktitle={The Thirteenth International Conference on Learning Representations},
year={2025},
url={https://openreview.net/forum?id=Iyrtb9EJBp}
}

@inproceedings{xie2022explanation,
  title={An Explanation of In-context Learning as Implicit Bayesian Inference},
  author={Xie, Sang Michael and Raghunathan, Aditi and Liang, Percy and Ma, Tengyu},
  booktitle={International Conference on Learning Representations},
  year={2022}
}

@misc{CVEProject2023cvelistV5,
  author       = {{CVE Project}},
  title        = {{CVE List V5}: {CVE} Cache of the Official {CVE} List in {CVE} {JSON} 5 Format},
  year         = {2023},
  publisher    = {GitHub},
  howpublished = {\url{https://github.com/CVEProject/cvelistV5}},
  note         = {Accessed: 2025-10-30}
}

@article{brown2020language,
  title={Language models are few-shot learners},
  author={Brown, Tom and Mann, Benjamin and Ryder, Nick and Subbiah, Melanie and Kaplan, Jared D and Dhariwal, Prafulla and Neelakantan, Arvind and Shyam, Pranav and Sastry, Girish and Askell, Amanda and others},
  journal={Advances in neural information processing systems},
  volume={33},
  pages={1877--1901},
  year={2020}
}

@inproceedings{min2022rethinking,
  title={Rethinking the role of demonstrations: What makes in-context learning work?},
  author={Min, Sewon and Lyu, Xinxi and Holtzman, Ari and Artetxe, Mikel and Lewis, Mike and Hajishirzi, Hannaneh and Zettlemoyer, Luke},
  booktitle={Proceedings of the 2022 conference on empirical methods in natural language processing},
  pages={11048--11064},
  year={2022}
}
